\documentclass{llncs}

\usepackage{makeidx}
\usepackage[utf8]{inputenc}
\usepackage{amsmath,amsfonts,bm,bbm,algorithm,algorithmic,amssymb}
\usepackage{hyperref,graphicx,sidecap}
\usepackage{algorithm,algorithmic}
\usepackage{tikz}
\usepackage{todonotes} 
\usetikzlibrary{bayesnet}

\usepackage{float}

\usepackage{graphicx}
\usepackage{subcaption}

\usepackage{enumitem}
\setlist{nolistsep}

\usepackage{color}

\definecolor{pcol}{HTML}{006600}
\definecolor{lcol}{HTML}{0033CC}

\begin{document}

\setlength{\abovedisplayskip}{7pt}
\setlength{\belowdisplayskip}{7pt}

\title{The Polylingual Labeled Topic Model}

\author{Lisa Posch\inst{1,2} \and Arnim Bleier\inst{1} \and Philipp Schaer\inst{1} \and Markus Strohmaier\inst{1,2}}

\institute{GESIS -- Leibniz Institute for the Social Sciences\\
Cologne, Germany
\and
Institute for Web Science and Technologies\\
University of Koblenz-Landau, Germany \\
\email{\{firstname.lastname\}@gesis.org}}

\date{\today}
\maketitle

\begin{abstract}
In this paper, we present the \emph{Polylingual Labeled Topic Model}, a model which combines the characteristics of the existing \emph{Polylingual Topic Model} and \emph{Labeled LDA}. The model accounts for multiple languages with separate topic distributions for each language while restricting the permitted topics of a document to a set of predefined labels. 
We explore the properties of the model in a two-language setting on a dataset from the social science domain. Our experiments show that our model outperforms LDA and Labeled LDA in terms of their held-out perplexity and that it produces semantically coherent topics which are well interpretable by human subjects.
\end{abstract}


\section{Introduction}

Topic models are a popular and widely used method for the analysis of textual corpora. \emph{Latent Dirichlet Allocation (LDA)} \cite{blei2003latent}, one of the most popular topic models, has been adapted to a multitude of different problem settings, such as modeling labeled documents with \emph{Labeled LDA (L-LDA)} \cite{ramage2009labeled} or modeling multilingual documents with \emph{Polylingual Topic Models (PLTM)} \cite{mimno2009polylingual}. Textual corpora often exhibit both of these characteristics, containing documents in multiple languages which are also annotated with a classification system.
However, there is currently no topic model which possesses the ability to process multiple languages while simultaneously incorporating the documents' labels. 

To close this gap, this paper introduces the \emph{Polylingual Labeled Topic Model (PLL-TM)}, a model which combines the characteristics of PLTM and L-LDA. \mbox{PLL-TM} models multilingual labeled documents by generating separate distributions over the vocabulary of each language, while restricting the permitted topics of a document to a set of predefined labels. 
We explore the characteristics of our model in a two-language setting, with German natural language text as the first language and the controlled \emph{SKOS} vocabulary of a thesaurus as the second language. The labels of the documents, in our setting, are classes from the classification system with which our corpus is annotated. 

\noindent\textbf{Contributions.} The main contribution of this paper is the presentation of the PLL-TM. We present the model's generative storyline as well as an easy-to-implement
inference strategy based on Gibbs sampling. For evaluation, we compute the held-out perplexity and conduct a word intrusion task with human subjects using a dataset from the social science domain. On this dataset, the PLL-TM outperforms LDA and L-LDA in terms of its predictive performance and generates semantically coherent topics. To the best of our knowledge, PLL-TM is the first model which accounts for multiple vocabularies and, at the same time, possesses the ability to restrict the topics of a document to its labels.

\section{Related Work}
\label{related_work}

Topic models are generative probabilistic models for discovering latent topics in documents and other discrete data. 
One of the most popular topic models, LDA, is a generative Bayesian model which was introduced by Blei et al. \cite{blei2003latent}. In this section, we review LDA, as well as the two other topic models whose characteristics we are going to integrate into PLL-TM.

\noindent \textbf{LDA.} Beginning with LDA \cite{blei2003latent}, we follow the common notation of a document $d$ being a vector of $N_d$ words, $\bm{w}_d$, where each word $w_{di}$ is chosen from a vocabulary of $V$ terms. A collection of documents is defined by \thinmuskip=0mu{$\mathcal{D} = \{\bm{w}_1,...,\bm{w}_D\}$}. LDA's generative storyline can be described by the following steps. 
\begin{enumerate}
\item 
For each document $d \in \{1,...,D\}$, a distribution $\theta_d$ over topics is drawn from a symmetric K-dimensional Dirichlet prior parametrized by $\alpha$:
\begin{align}
    \theta_d \sim Dir(\alpha) \mbox{ .}
\end{align} 
\item 
Then, for each topic $k = \{1,...,K\}$, a distribution $\phi_{k}$ over the vocabulary is drawn form a V-dimensional Dirichlet distribution parametrized by $\beta$:
\begin{align}
    \phi_{k} \sim Dir(\beta) \mbox{ .}
\end{align}
\item 
In the final step, the $i^{th}$ word in document $d$ is generated by first drawing a topic index $z_{di}$ and subsequently, a word $w_{di}$ from the topic indexed by $z_{di}$:
\begin{align}
    w_{di} \sim Cat(\phi_{z_{di}}) \mbox{ ,} &&& z_{di} \sim Cat(\theta_d)\mbox{ .}
\end{align}
\end{enumerate}

\noindent \textbf{Labeled LDA.} 
Ramage et al. \cite{ramage2009labeled} introduced L-LDA, a supervised version of LDA. In L-LDA, a document $d$'s topic distribution $\theta_d$ is restricted to a subset of all possible topics $\bm{\Lambda}_d \subseteq \{1,..,K\}$. 
Here, collection of documents is defined by $\mathcal{D} = \{(\bm{w}_1, \bm{\Lambda}_1),...,(\bm{w}_D, \bm{\Lambda}_D)\}$.
The first step in L-LDA's generative storyline draws the distribution of topics  $\theta_d$ for each document $d \in \{1,...,D\}$
\begin{align}
    \theta_d \sim Dir(\alpha \bm{\mu}_d) \mbox{ ,}
\end{align}
where $\alpha$ is a continuous positive valued scalar and $\bm{\mu}_d$ is a K-dimensional vector
\begin{align}
    \label{eq:label_vec}
    \mu_{dk} =
    	\begin{cases} 
		1 & \text{if } k \in \bm{\Lambda}_d  \\
		0 & \text{otherwise} \mbox{ ,}
	\end{cases} 
\end{align}
indicating which topics are permitted. Once these label-restricted topic distributions are drawn, the process of generating documents continues identically to the generative process of LDA. In the case of $\bm{\Lambda}_d = \{1,..,K\}$ for all documents, no restrictions are active and L-LDA is reduced to LDA.

\noindent \textbf{Polylingual Topic Model.}  Ni et al. \cite{ni2009mining} extended the generative view of LDA to multilingual documents.  Mimno et al. \cite{mimno2009polylingual} elaborated on this concept, introducing the \textit{Polylingual Topic Model} (PLTM). PLTM assumes that the documents are available in $L$ languages. A document $d$ is represented by $[\bm{w}_d^1,...,\bm{w}_d^L]$, where for each language $l \in {1,...,L}$, the vector $\bm{w}_d^l$ consists of $N_{d}^l$ words which are chosen from a language specific vocabulary with $V^l$ terms. A collection of documents is then defined by $\mathcal{D} = \{[\bm{w}_1^1,...,\bm{w}_1^L],...,[\bm{w}_D^1,...,\bm{w}_D^L]\}$. The generative storyline is equivalent to LDA's except that steps 2 and 3 are repeated for each language. Hence, for each topic $k = \{1,...,K\}$ in each language $l \in \{1,...,L\}$, a language specific topic distribution $\phi_{k}^l$ over the vocabulary of length $V^l$ is drawn:
\begin{align}
    \phi_{k}^l \sim Dir(\beta^l) \mbox{ .}
\end{align}
Then, the $i^{th}$ word of language $l$ in document $d$ is generated by drawing a topic index $z_{di}^l$ and subsequently, a word $w_{di}^l$ from a language specific topic distribution indexed by $z_{di}^l$:
\begin{align}
    w_{di}^l \sim Cat(\phi^l_{z^l_{di}}) \mbox{ ,} &&& z_{di}^l \sim Cat(\theta_d)  \mbox{ .}
\end{align}
Note that in the special case of just one language, i.e. $L=1$, PLTM is reduced to LDA.

 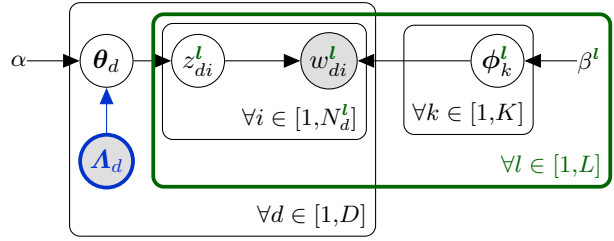
\begin{figure}[t]
	\centering
	\begin{center}
\begin{tikzpicture}

  \node[latent]				(z) {$z_{di}^{\textcolor{pcol}{\bm{l}}}$};
  \node[obs, right=of z]		(w) {$w_{di}^{\textcolor{pcol}{\bm{l}}}$};
  \node[latent, right=1.5cm of w]      	(phi) {$\bm{\phi}_k^{\textcolor{pcol}{\bm{l}}}$};
  \node[const, right=.7cm of phi]	(beta) {$\beta^{\textcolor{pcol}{\bm{l}}}$};
  \node[latent, left=.5 of z]		(theta) {$\bm{\theta}_d$};  
  \node[obs, below=0.6cm of theta, draw=lcol, line width=.5mm, text=lcol]	(lambda) {$\bm{\Lambda}_d$};
  \node[const, left=.7cm of theta]	(alpha) {$\alpha$};
  \coordinate[above=0.3cm of theta]	(hidden);
  
  \edge {z,phi} {w};
  \edge {beta} {phi}; 
  \edge {theta} {z};
  \edge[lcol] {lambda} {theta};
  \edge {alpha} {theta};  

  \plate {words} {(w)(z)} {$\forall \, i \in [1,N_{d}^{\textcolor{pcol}{\bm{l}}}]$} ;
  \plate {documents} {(words)(theta)(lambda)(hidden)} {$\forall \, d \in [1,D]$} ;
  \plate {topics} {(phi)} {$\forall \, k \in [1,K]$} ;
  \coloredplate {languages} {(words)(topics)(z)(w)(phi)(beta)} {$\forall \, l \in [1,L]$} ;

\end{tikzpicture}

	\end{center}	
  	\caption{\textbf{The PLL-TM in plate notation.} Random variables are represented by nodes. Shaded nodes denote the observed words and labels, bare symbols indicate the fixed priors $\alpha$ and $\beta^l$. Directed edges between the nodes then define conditional probabilities, where the child node is conditioned on its parents. The rectangular plates indicate replication over data-points and parameters. Colors indicate the parts which are inherited from L-LDA (\textbf{\textcolor{lcol}{blue}}) and PLTM (\textbf{\textcolor{pcol}{green}}). Black is used for the LDA base.}
  \centering
  \label{fig:plate_notation}
\end{figure}

\section{The Polylingual Labeled Topic Model}
\label{model}

In this section, we introduce the \emph{Polylingual Labeled Topic Model (PLL-TM)}, which integrates the characteristics of the models described in the previous section into a single model. 
Figure \ref{fig:plate_notation} depicts the PLL-TM in plate notation. Here, a collection of documents is defined by \thinmuskip=0mu{$\mathcal{D} = \{[\bm{w}_1^1,...,\bm{w}_1^L], \bm{\Lambda}_1)),...,[\bm{w}_D^1,...,\bm{w}_D^L], \bm{\Lambda}_D)\}$}.

The generative process follows three main steps:
\begin{enumerate}
\item For each document $d \in \{1,...,D\}$, we draw the distribution of topics
\begin{align}
    \theta_d \sim Dir(\alpha \bm{\mu}_d) \mbox{ ,}
\end{align}
where $\bm{\mu}_d$ is computed according to Equation \ref{eq:label_vec}. 

\item For each topic $k \in \{1,...,K\}$ in each language $l \in \{1,..., L\}$, we draw a distribution over the vocabulary of size $V^l$:
\begin{align}
    \phi_{k}^l \sim Dir(\beta^l) \mbox{ ,}
\end{align}
\item Next, for each word in each language $l$ of document $d$, we draw a topic 
\begin{align}
    w_{di}^l \sim Cat(\phi_{z_{di}^l}^l) \mbox{ ,} &&& z_{di}^l \sim Cat(\theta_d)\mbox{ .}
\end{align}
\end{enumerate}
Note that PLL-TM contains both PLTM and L-LDA as special cases.

For inference, we use collapsed Gibbs sampling \cite{griffiths2004finding} for the indicator variables $\bm{z}$, with all other variables integrated out. The full conditional probability for a topic $k$ is given by 
\begin{align}
    P(z_{di}^l = k \mid w_{di}^l = t, ...) \propto
    \dfrac{n_{dk}^{\neg di} +\alpha}{n_{d.}^{\neg di} + K\alpha} \times
    \dfrac{n_{kt}^{l\neg di} +\beta^l}{n_{k.}^{l\neg{di}} + V^l\beta^l} \mbox{ ,}
\end{align}
where $n_{dk}$ is the number of tokens allocated to topic $k$ in document $d$, and $n^l_{kt}$ is the number of tokens of word $w_{di}^l = t$ which are assigned to topic $k$ in language $l$. Furthermore, $\cdot$ is used in place of a variable to indicate that the sum over its values (i.e. $n_{d.} = \sum_k n_{dk}$ ) is taken and $\neg{di}$ to mark the current token as excluded. While the full conditional posterior distribution is reminiscent of the one used in PLTM, the assumptions of the L-LDA model restrict the probability $P(z_{di}^l = k)$ to those $k \in \bm{\Lambda}_d$ with which document $d$ is labeled.
\begin{table}[b]
\caption{This table shows the five most probable terms for two classes in the CSS, generated by PLL-TM, in two languages: $TheSoz$ (TS) and German natural language words with their translation (AB).}
\resizebox{0.9\textwidth}{!}{\begin{minipage}{1\textwidth}
\begin{tabular}{l l}
 &  \\[.05mm] 
  & \textbf{\small Population Studies, Sociology of Population:} \\
   \textit{TS:} & {\footnotesize \underline{\smash{\textit{population development}}}, \underline{\smash{\textit{demographic aging}}}, \underline{\smash{\textit{population}}}, \underline{\smash{\textit{demographic factors}}}, \underline{\smash{\textit{demography}}}}\\
   \textit{AB:} & 
   {\footnotesize wandel, demografischen, bevölkerung, deutschland, entwicklung} \\
   & {\footnotesize (change, demographic, population, germany, development)} \\
   & \footnotesize \\
   & \textbf{\small Developmental Psychology:}  \\
      \textit{TS:} & {\footnotesize \underline{\smash{\textit{child}}}, \underline{\smash{\textit{developmental psychology}}}, \underline{\smash{\textit{adolescent}}}, \underline{\smash{\textit{personality development}}}, \underline{\smash{\textit{socialization research}}}} \\
   \textit{AB:} & 
   {\footnotesize entwicklung, sozialisation, kinder, kindern, identität} \\
   & {\footnotesize (development, socialization, children, children, identity)}
\end{tabular}
\label{tab:topics}
\end{minipage} }
\end{table}

\section{Evaluation}
\label{evaluation}

\begin{figure}[t!]
        \centering
        \begin{subfigure}[b]{0.47\textwidth}
                \includegraphics[width=\textwidth]{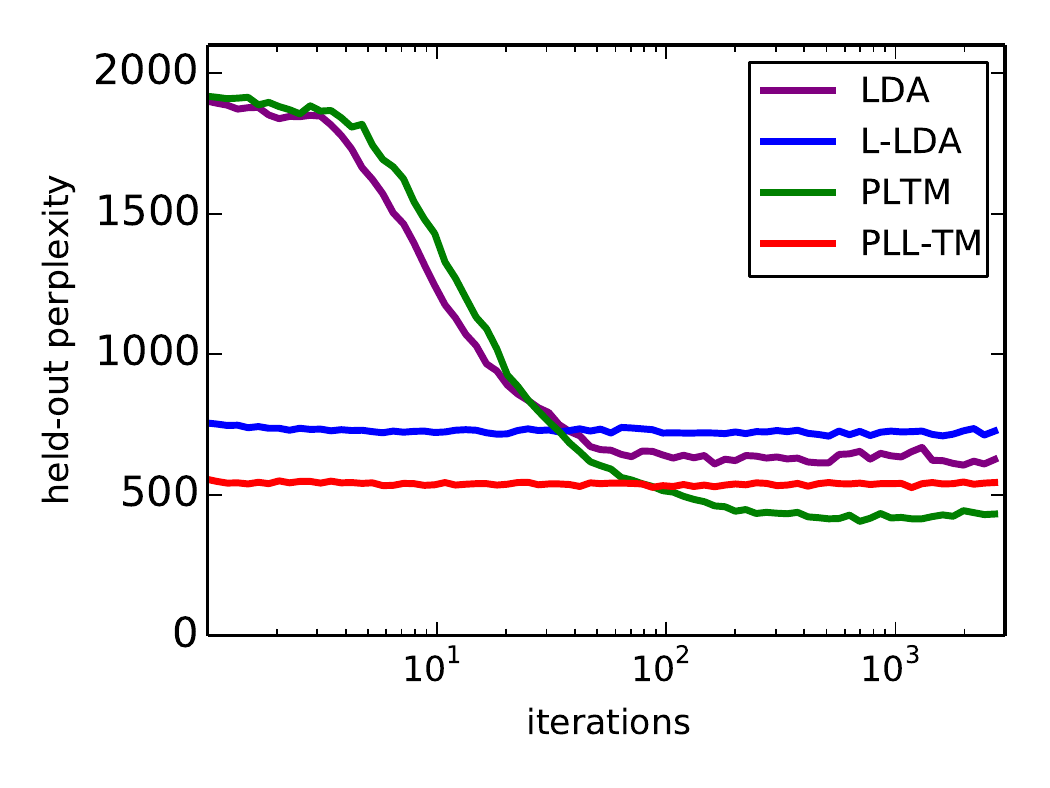}
                \caption{Comparison of the held-out perplexity (lower values are better) as a function of iterations.}
                \label{fig:hybrid_ppx}
        \end{subfigure}
        ~
        \begin{subfigure}[b]{0.47\textwidth}
                \includegraphics[width=\textwidth]{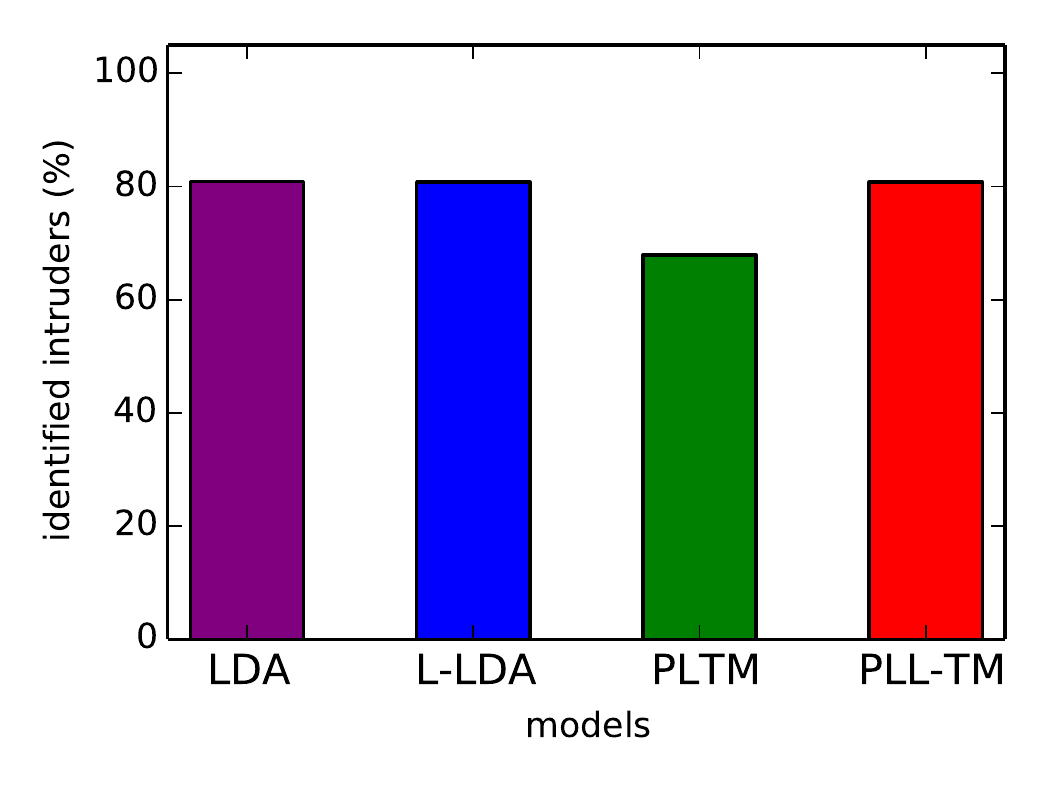}
                \caption{Comparison of the semantic coherence (word intrusion) of the generated topics.}
                \label{fig:crowdflower}
        \end{subfigure}
				\caption{\textbf{Evaluation of the PLL-TM.} These figures show that on the SOLIS dataset, PLL-TM outperforms LDA and L-LDA in terms of its predictive performance and produces topics with a higher semantic coherence than PLTM.}
				\label{fig:results}
\end{figure}

For our evaluation, we use documents from the \emph{Social Science Literature Information System (SOLIS)}. The documents are manually indexed with the SKOS \emph{Thesaurus for the Social Sciences (TheSoz)} \cite{zapilko2013thesoz} and manually classified with the \emph{Classification for the Social Sciences (CSS)} by human domain experts. For our experiments, we used all SOLIS documents which were published in the years 2008 to 2013, resulting in a corpus of about 60.000 documents.

We explore the characteristics of our model in a two-language setting, with German natural language text as the first language ($AbstractWords$) and the controlled \emph{SKOS} vocabulary of a thesaurus as the second language ($TheSoz$). The labels of the documents, in our setting, are classes from the CSS. 
After applying standard preprocessing to remove rare words and stopwords, $TheSoz$ consisted of 802.764 tokens over a vocabulary of 7.406 distinct terms, and $Abstract\-Words$ consisted of 5.417.779 tokens over a vocabulary of about 43.000 distinct terms. In our corpus, each document is labeled with an average of 2.14 classes. 

We compare four different topic models: LDA, L-LDA, PLTM and PLL-TM. The unilingual models (i.e. LDA and L-LDA) were trained on language $TheSoz$; the polylingual models (i.e. PLTM and PLL-TM) were trained on $TheSoz$ and $AbstractWords$. The documents in our corpus were labeled with a total of 131 different classes from the CSS and we trained the unlabeled models with an equal number of topics. $\alpha$ and $\beta^l$ were specified with 0.1 and 0.01, respectively.  Table \ref{tab:topics} shows the topics generated by PLL-TM for two classes of the CSS, reporting the five most probable terms for the languages $TheSoz$ and $Abstract\-Words$.

\textbf{Language Model Evaluation.} For an evaluation of the predictive performance, we computed the held-out perplexity for all models. We held out 1.000 documents as test set $\mathcal{D}_{test}$ and, with the remaining data $\mathcal{D}_{train}$, we trained the four models. 
We split each test document in the following way:
\begin{itemize}
\item $\bm{x}_{d1}$: All words of language $AbstractWords$ and a randomly selected 50\% of the words in language $TheSoz$ which occur in document $d$.
\item $\bm{x}_{d2}$: The remaining 50\% of the words in language $TheSoz$ which occur in document $d$.
\end{itemize}
The test documents for the unilingual models were split analogously, with $\bm{x}_{d1}$ consisting of 50\% percent of the words in language $TheSoz$ which occur in document $d$. For each document $d$, we computed the perplexity of $\bm{x}_{d2}$. 

Figure \ref{fig:hybrid_ppx} shows the results of this evaluation. One can see that the labeled models both start out with a lower perplexity and need less iterations to achieve a good performance, which is due to the fact that the labels provide additional information to the model. In contrast, the unlabeled models need almost 100 iterations to achieve a comparable performance. On our corpus, PLL-TM outperformed LDA and L-LDA, and even though PLL-TM had a higher perplexity than PLTM, it is important to keep in mind that PLTM does not possess the ability to produce topics which correspond to the classes of the CSS.

\textbf{Human Evaluation of the Topics.} 
Chang et al. \cite{chang2009reading} proposed a formal setting in which humans evaluate the latent space of a topic model. For evaluating the topics' semantic coherence, they proposed a \emph{word intrusion} task: Crowdworkers were shown six terms, five of which were highly probable terms in a topic and one was an ``intruder" -- an improbable term for this topic which had a high probability in some other topic. 

We conducted the word intrusion task for the four topic models on CrowdFlower \cite{biewald2012massive}, with ten distinct workers for each topic in each model. 
Figure \ref{fig:crowdflower} shows the results of this evaluation for the different models. For each model, the figure depicts the percentage of topics for which the ten workers collectively detected the correct intruder. The collective decision was based on CrowdFlower's \emph{confidence score}, i.e. the level of agreement between workers weighted by each worker's percentage of correctly answered test questions.
The results show that PLL-TM produces topics which are equally coherent as unilingual models, and more coherent than the topics produced by PLTM.

\section{Discussion and Conclusions}
\label{conclusion}

In this paper, we presented PLL-TM, a joint model for multilingual labeled documents. The results of our evaluation showed that PLL-TM was the only model which produced both highly interpretable topics and achieved a good predictive performance. Compared to L-LDA, the only other model capable of incorporating label information, our model produced equally well interpretable topics while achieving a better predictive performance. Compared to PLTM, the only other model capable of dealing with multiple languages, PLL-TM had a lower predictive performance, but produced topics with a higher semantic coherence.
For future work, we plan an evaluation of the model in a label prediction task and an application of the model in a setting with more than two natural languages.
Furthermore, we plan an evaluation on a larger dataset using a more memory-friendly inference strategy such as \textit{Stochastic Collapsed Variational Bayesian Inference} \cite{foulds2013stochastic}, which has been shown to be applicable outside of its original LDA application \cite{bleier2013}.

\bibliographystyle{abbrv}
\bibliography{bibliography}

\begin{thebibliography}{10}

\bibitem{biewald2012massive}
L.~Biewald.
\newblock Massive multiplayer human computation for fun, money, and survival.
\newblock In {\em Current Trends in Web Engineering - Workshops, Doctoral
  Symposium, and Tutorials, Held at {ICWE} 2011, Paphos, Cyprus, June 20-21,
  2011. Revised Selected Papers}, pages 171--176, 2011.

\bibitem{blei2003latent}
D.~M. Blei, A.~Y. Ng, and M.~I. Jordan.
\newblock Latent dirichlet allocation.
\newblock {\em Journal of Machine Learning Research}, 3:993--1022, 2003.

\bibitem{bleier2013}
A.~Bleier.
\newblock Practical collapsed stochastic variational inference for the hdp.
\newblock In {\em NIPS Workshop on Topic Models: Computation, Application, and
  Evaluation}, 2013.

\bibitem{chang2009reading}
J.~Chang, J.~L. Boyd{-}Graber, S.~Gerrish, C.~Wang, and D.~M. Blei.
\newblock Reading tea leaves: How humans interpret topic models.
\newblock In {\em Advances in Neural Information Processing Systems 22: 23rd
  Annual Conference on Neural Information Processing Systems 2009. Proceedings
  of a meeting held 7-10 December 2009, Vancouver, British Columbia, Canada.},
  pages 288--296, 2009.

\bibitem{foulds2013stochastic}
J.~R. Foulds, L.~Boyles, C.~DuBois, P.~Smyth, and M.~Welling.
\newblock Stochastic collapsed variational bayesian inference for latent
  dirichlet allocation.
\newblock In {\em The 19th {ACM} {SIGKDD} International Conference on Knowledge
  Discovery and Data Mining, {KDD} 2013, Chicago, IL, USA, August 11-14, 2013},
  pages 446--454, 2013.

\bibitem{griffiths2004finding}
T.~L. Griffiths and M.~Steyvers.
\newblock Finding scientific topics.
\newblock {\em Proceedings of the National Academy of Sciences}, 2004.

\bibitem{mimno2009polylingual}
D.~M. Mimno, H.~M. Wallach, J.~Naradowsky, D.~A. Smith, and A.~McCallum.
\newblock Polylingual topic models.
\newblock In {\em Proceedings of the 2009 Conference on Empirical Methods in
  Natural Language Processing, {EMNLP} 2009, 6-7 August 2009, Singapore, {A}
  meeting of SIGDAT, a Special Interest Group of the {ACL}}, pages 880--889,
  2009.

\bibitem{ni2009mining}
X.~Ni, J.~Sun, J.~Hu, and Z.~Chen.
\newblock Mining multilingual topics from wikipedia.
\newblock In {\em Proceedings of the 18th International Conference on World
  Wide Web, {WWW} 2009, Madrid, Spain, April 20-24, 2009}, pages 1155--1156,
  2009.

\bibitem{ramage2009labeled}
D.~Ramage, D.~L.~W. Hall, R.~Nallapati, and C.~D. Manning.
\newblock Labeled {LDA:} {A} supervised topic model for credit attribution in
  multi-labeled corpora.
\newblock In {\em Proceedings of the 2009 Conference on Empirical Methods in
  Natural Language Processing, {EMNLP} 2009, 6-7 August 2009, Singapore, {A}
  meeting of SIGDAT, a Special Interest Group of the {ACL}}, pages 248--256,
  2009.

\bibitem{zapilko2013thesoz}
B.~Zapilko, J.~Schaible, P.~Mayr, and B.~Mathiak.
\newblock Thesoz: {A} {SKOS} representation of the thesaurus for the social
  sciences.
\newblock {\em Semantic Web}, 4(3):257--263, 2013.

\end{thebibliography}

\end{document}